\documentclass[11pt]{article}

\newcommand{\mdiv}{\nabla\cdot}
\newcommand{\mvec}[1]{{\bf #1}}
\newcommand{\mhvec}[1]{{\bm{\hat #1}}}
\newcommand{\mcurl}{\nabla\times}
\newcommand{\nabb}{\nabla^2}

\newcommand{\theosea}{{\em TheoSea}}

\newcommand{\fld}{{\cal F}}

\usepackage{graphicx}
\usepackage{color}
\usepackage{amsmath}
\usepackage{bm}

\title{\theosea: Marching Theory to Light}

\author{Mark A.~Stalzer\footnote{Center for Data-Driven Discovery, California Institute of Technology, stalzer at caltech.edu}\ \ and
Chao Ju\footnote{Minerva Schools at KGI, jj at minerva.kgi.edu}
}

\begin{document}

\maketitle

\begin{abstract}
There is sufficient information in the far-field of a radiating dipole antenna to rediscover the Maxwell Equations and the wave equations of light, including the speed of light $c.$ \theosea\ is a Julia program that does this in about a second, and the key insight is that the {\em compactness} of theories drives the search. The program is a computational embodiment of the scientific method: observation, consideration of candidate theories, and validation.
\end{abstract}

\section{Motivation and Background}
  This work flowed from a comment in the concluding remarks of a recent review (2016) of work in data-driven scientific discovery\cite{stalzer16}. Specifically,
\begin{quote}
  {\em \ldots it may be within current computing and algorithmic technology to infer the Maxwell Equations directly from data given knowledge of vector calculus.}
\end{quote}
This paper reports on recent progress towards this objective.

The overarching goal is to develop methods that can infer compact theories from data. Most data intensive analysis techniques are based on machine learning or statistics. These are quite useful, but do not lead to deep understanding or insight. The scientific method and creative scientists have been very good at observations (experiments) and building human understandable models (theory). In this program, we turn both of these ideas on their heads: can a computer given an appropriate {\em virtual experiment} (VE) figure out mathematically compact theories? The initial applications are in electrodynamics (the Maxwell Equations)\cite{maxwell65}, and there are many other examples such as thermodynamics. Eventually, it is hoped the methods developed will be applicable to data sets from real measurements in a wide variety of fields in physics, engineering, and economics.

\paragraph{The Maxwell Equations.}
The Maxwell Equations in free space with the transformation $\mvec{B'} = c\,\mvec{B}$ are:
\begin{eqnarray}
  \mdiv\mvec{E} & = & 0 \label{eqn:maxEdiv} \\
  \mdiv\mvec{B'} & = & 0 \label{eqn:maxBdiv} \\
  \mcurl\mvec{E} + {1 \over c}\, {\partial \mvec{B'} \over {\partial t}} & = & 0 \label{eqn:maxEcurl} \\
  c\,\mcurl\mvec{B'} - {\partial\mvec{E} \over {\partial t}} & = & 0\label{eqn:maxBcurl}
\end{eqnarray}
where $c = 2.99792458\times 10^{8} m/s$ (MKS units). The spatial-temporal coupling of $\mvec{E}$ and $\mvec{B}$ is how we get electromagnetic waves. The utitility of the $\mvec{B}$ transformation for numerical stability is discussed in Sec.~\ref{sec:validation} and that is why the Equations look in a slightly strange form in terms of constants.

\paragraph{Problem, basic approach, and plan.} The problem is to computationally rediscover the Maxwell Equations from data. The data consists of a set of virtual experiments as described in Sec.~\ref{sec:virt}. The experiments are simulated far-field measurements from a dipole antenna. In principle, real data could be used but it is easier to do this purely computationally. This is discussed more in the concluding remarks in Sec.~\ref{sec:remarks}.

The second step is the generation of candidate theories in Sec.~\ref{sec:theosea}, and the third is validation in Sec.~\ref{sec:validation}. Validation is what connects observations to candidate theories and this is the essence of fact based scientific discovery. Here it is done with linear algebra. The final results are in Sec.~\ref{sec:results}, particularly Fig.~\ref{fig:out}.

But before going into the details, a few comments about past work and Julia.

\paragraph{Past work.} Attempts to use computers to rediscover physical laws goes back to at least 1979 with BACON.3\cite{langley79}. The program successfully found the ideal gas law, $PV = nRT,$ from small data tables.\footnote{Perhaps the NFL should have consulted BACON.} One of us (Stalzer) and William Xu of Caltech have also rediscovered the ideal gas law with Van der Waals forces using the approach of this paper\cite{xu17}. In 2009, researchers rediscovered the kinematic equation for the double pendulum essentially using optimization methods to fit constants to candidate equations\cite{schmidt09}.

What differentiates this work is twofold: the concept of search driven by compactness and completeness, and targeting electrodynamics which is mathematically a much more difficult theory. Indeed, electrodynamics was the first unification (the electric and magnetic fields), and Einstein's special relativity is baked right into the equations once the brilliant observation is made that $c$ is the same in all inertial reference frames. \theosea\ also finds the wave equation of light as a consequence of the rediscovered free space Maxwell Equations.

\paragraph{Julia.} \theosea\ is written in Julia\cite{julia17}, a relatively recent language (roughly 2012) that is both easy to use and has high performance. Julia can be programmed at a high expressive level, and yet given enough type information it automatically generates efficient machine code. \theosea\ is a Julia {\em meta-program} that writes candidate theories in terms of Julia sets that are then validated against data. The set elements are compiled Julia expressions corresponding to terms in the candidate theories.

\section{Observations and the Virtual Experiment}
\label{sec:virt}

\begin{figure}[th]
\centering
\includegraphics[width=2.5truein]{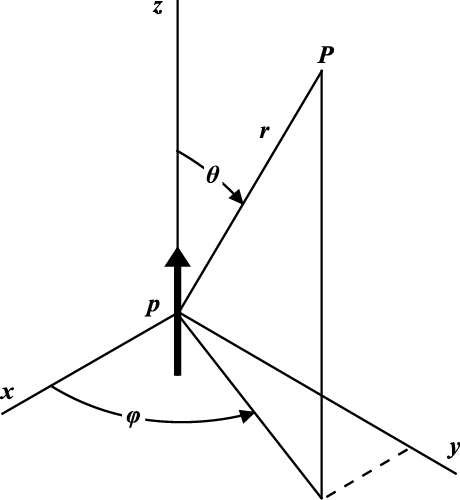}
\caption{Geometry of the far-field of a dipole with moment ${\bf p}$ at the origin oscillating at angular frequency $\omega$ at a far point $P.$}
\label{fig:dipole}
\end{figure}

The data is from the far-field of a radiating antenna for $\mvec{E}, \mvec{B}$ as shown in the geometry Fig.~\ref{fig:dipole} and data Tab.~\ref{tab:data}.

\begin{table}[th]
\centering
\begin{tabular}{lll|ll}
\hline
$r\,[10^{17}]$ & $\phi$ & $\theta$ & $\mvec{E_P}(0)\,[10^{-16}]$ & $\mvec{B_P}(0)\,[10^{-16}]$ \\
\hline
1 & 0.905 & 0.997 & (2.713,3.455,-6.793) & (-6.363,4.996, 0) \\
 1 & 2.767 & 2.908 & (2.018,-0.794,-0.516) & (-0.816,-2.074, 0) \\
 1 & 4.631 & 0.291 & (-0.214,-2.639,-0.793) & (2.754,-0.224, 0) \\
 1 & 5.597 & 3.051 & (-0.666,0.546,-0.078) & (0.548,0.669, 0) \\
  1 & 0.468 & 2.369 & (-4.299,-2.170,-4.690) & (-3.029,6.001, 0) \\
\hline
\end{tabular}
\caption{Experiments with geometry parameters and the corresponding field observations at point $P$ and time $t;$ adjusted for scale.}
\label{tab:data}
\end{table}

The fields at a far point $P$ are\cite{griffiths15}:
\begin{eqnarray}
\textbf{E} & = & -\frac{\mu_0 p_0 \omega^2}{4\pi} (\frac{\sin\theta}{r})\cos{[\omega(t-r/c)]}{\mhvec{\theta}} \\
\textbf{B} & = & -\frac{\mu_0 p_0 \omega^2}{4\pi c}(\frac{\sin\theta}{r})\cos{[\omega(t-r/c)]}{\mhvec{\phi}}
\end{eqnarray}
where $\mu_0 = 4 \pi \times 10^{-7}$ is the permeability of free space, $p_0$ is the strength of the dipole, and $\omega$ is the frequency of the dipole oscillation.

Five virtual experiments were done with various parameters $r, \phi, \theta$ with a fixed $\omega.$ The observables are $\mvec{E}(\mvec{x}, t)$ and $\mvec{B}(\mvec{x}, t),$ where $\mvec{x}$ is in the region of the point $P.$ The nice thing about this VE is that various space-time derivatives can be computed analytically. These experiments are show in Tab.~\ref{tab:data}, with the fields given at a steady state $t = 0.$

Rediscovering the Equations from this data is the topic of the two next sections.

\section{Rapid Enumeration of Candidate Theories}
\label{sec:theosea}

Given an alphabet ${\cal A}$ of symbols, such as operators and fields, a language ${\cal L}$ is {\em recursively enumerable} if there exists a Turing machine that will enumerate all valid strings in the language\cite{denning78}.

By the {\em infinite monkey theorem}\cite{borel13} the solution can be found --- if the constants are limited to rationals --- just by enumeration and validation. The goal of this section is to show a way of doing this enumeration in a tractable way that also finds the most compact theory.

\subsection{Abstract Enumeration}

Abstractly, think of an alphabet ${\cal A} = [A, B, C, \ldots]$ where any letter can appear once in a sentence and the length of the alphabet is $n.$ This is a simple combinatorial enumeration problem and the solution to the number of sets of size $m$ (later $m$ will be relabeled $q$) taken from ${\cal A}$ is $C(n, m).$ However, what if the symbols --- letters --- in the alphabet have different weights? What if the alphabet is more like ${\cal A} = [A=1, B=1, C=4, D=4, E=4, F=4, G=7, H=7, I=7, J=7, K=4,  L=7].$ This can dramatically decrease the enumeration size as shown in the next section, and the underlying motivation is shown in Sec.~\ref{sec:phys}.

The algorithm enumerates sets of increasing complexity $q,$ where $q$ is the sum of the alphabet letter weights in a given candiate theory. It can be thought of as a form Depth-First Iterative Deepening (DFID)\cite{korf85} first formalized by R.~E.~Korf in 1985. Optimality flows from a theorem by Korf:

\newtheorem{theo}{Theorem}

\begin{theo}[Korf 4.2]
Depth-first iterative-deepening is asymptotically optimal among brute-force tree searches in terms of time, space, and length of solution.
\end{theo}

By length of solution, Korf means the depth of the search where a solution is found. For \theosea\  compactness is the sum of the symbol weights along a potential solution branch in the search; as will be shown in Sec.~\ref{sec:phys}.

It is perhaps easiest to think of the algorithm inductively. There is a data structure {\tt theos} that holds all theorems (sets) of length $q$ and it is built up from $q = 1.$ The base cases are the singleton theories of a given complexity, so for the alphabet ${\cal A}$ we have {\tt theos[1] = [A, B]} and {\tt theos[4] = [D, ...]} and so on. So the base cases, such as $q=1$ are all set; and then for $q>1$ we use a $q: l,m$ ``Squeeze''. At step $q$ consider all theories that can possibly be of length $q,$ {\em marching} $l$ upward from 1 and $m$ downward from $q-1$ in a kind of double iteration. The correctness is immediate by Korf 4.2 and the fact that $q = l+m,$ too short theories are discarded $(<q)$, and set elements are unique. The Julia code is in the Appendix.

\begin{figure}[t]
\centering
\includegraphics[scale=0.8]{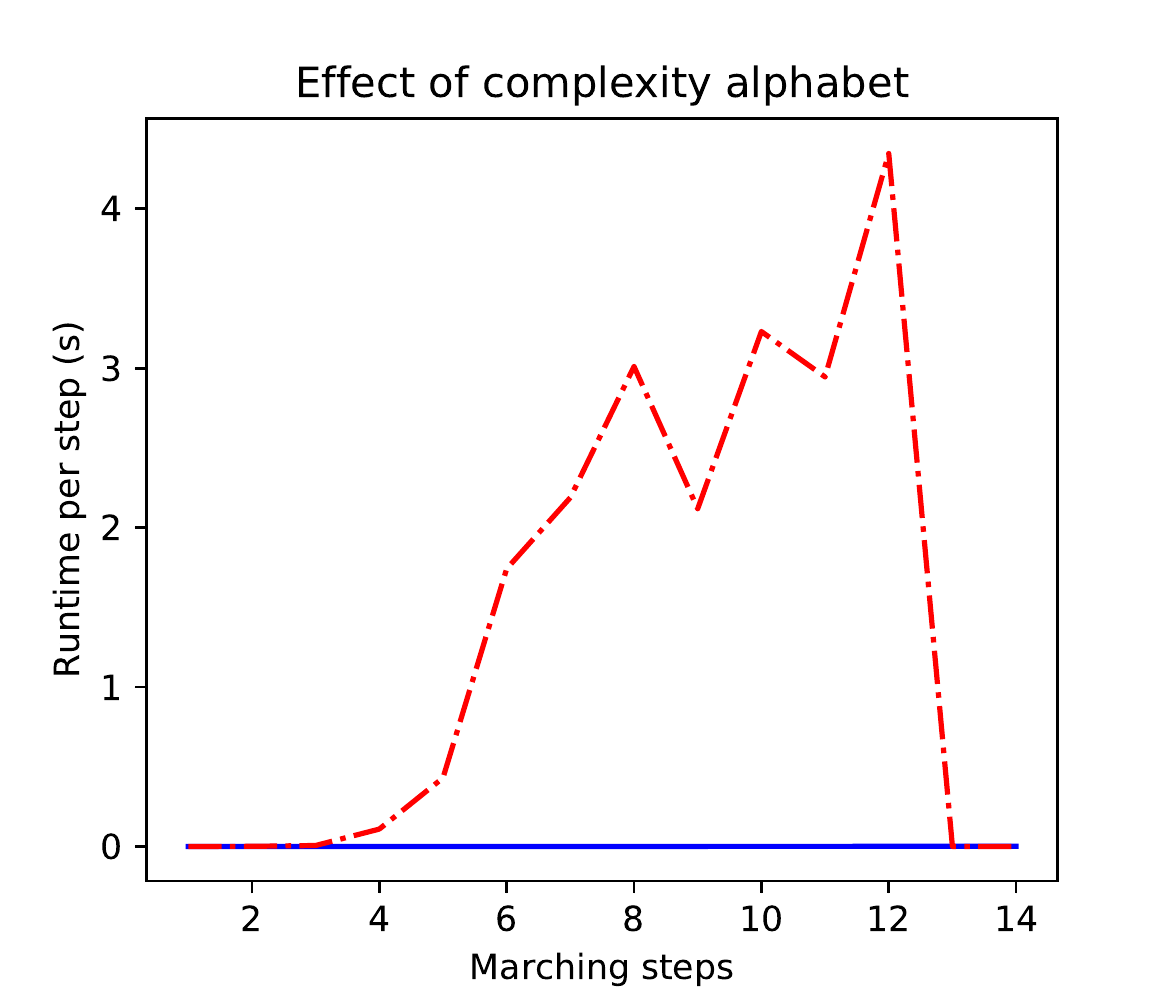}
\caption{Time to discovery (s): Fast versus Slow brute force. The Slow search cuts off at 12 due to maximum complexity.}
\label{fig:power}
\end{figure}

\paragraph{Performance.} The total times are ${\rm Fast} =0.006s$ for the weighted ${\cal A}$ above, and ${\rm Slow} =20.1s$ where the symbol weights are unity\footnote{The machine was a MacBook Pro (Retina, 13-inch, Late 2013) running a 2.8 GHz Intel Core i7 single-threaded using Julia 0.5.}. A graph is in Fig.~\ref{fig:power}: {\em compactness matters.}

\subsection{Relation to the Equations}
\label{sec:phys}
The underlying motivation was described above and here is the decoder ring; think of $A = \mvec{E}, B = \mvec{B}, \dots$

\begin{table}[h]
\centering
\begin{tabular}{lll}
\hline
Operator & Term Cost & Alphabet \\
\hline
$\fld$ & 1 & A, B \\
$\mdiv\fld$ & 4 & C, D\\
$\mcurl\fld$ & 7 & G, H\\
$\nabb\fld$ & 7 & I, J\\
 ${\partial \over \partial t}\fld$ & 4 & E, F\\
${\partial^2 \over \partial t^2}\fld$ & 7 & K, L\\
\hline
\end{tabular}
\caption{Complexity of each operator term working on a field $\fld \in \{\mvec{E}, \mvec{B}\}.$}
\label{tab:tcmplx}
\end{table}

The Maxwell Equations, up to constants, are $[C], [D],$ the $\mvec{E}, \mvec{B}$ divergence equations, both of complexity 4; and the field coupling equations of $[G, F]$ and $[H, E],$ of complexity 11. Also, the wave equations of light are $[I, K]$ and $[J, L],$ each of complexity 14\cite{fitzpatrick08}. {\em The complexity metric is just $1 +$ the number of space-time derivatives taken.} For the Maxwell Equations, the total discovery time with about $1.1s$ as is reported in Sec.~\ref{sec:results}.

\section{Theory Validation: Fitting Constants}
\label{sec:validation}

The glue that connects the enumeration (Sec.~\ref{sec:theosea}) to the virtual experiments (Sec.~\ref{sec:virt}) is finding constants in the candidate theories that fit the data: {\em if the do not exist, which is almost always the case, the theory is invalid. However, if they do exist then the theory is valid with high probability.} This is a linear algebra problem as described below.

Finding the constants is equivalent to finding the null space of a linear system (the data matrix extracted from Tab.~\ref{tab:data}). If the dimension of the null space is 0, then the theory is not valid because the only solution is a trivial zero vector. If the dimension of the null space is nonzero, it can only be 1, which corresponds to a unique solution. The reason is that in our enumerative method, we remove all valid sub-theories from the candidate theory before determining its constants. Had the dimension of the null space of the resulting system be larger than 1, it would have implied that some sub-theory is not removed, contradicting the hypothesis. Next, to find the rank of the null space and the null space itself, we cannot simply use Julia's built-in {\tt rank()} or {\tt nullspace()} functions because the dynamic ranges are large $(> 10^{30}).$\footnote{This is one of those cases when symbolics and numerics do not play well together.}

The solution is to use singular value decomposition, in which the number of zero singular values (SVs) is equivalent to the dimension of the null space. The insight is that if we scale $\mvec{B}$ by a factor of $c,$ it will be on the same scale as $\mvec{E},$ and after normalizing each column of the matrix so that each is on the same scale with another, the resulting singular values (if nonzero) should also be on the same scale. After scaling and normalizing, we use Julia's built-in {\tt svdvals()} function to obtain a list of SVs ranked from the largest to the smallest.

As discussed previously, the dimension of the null space can either be 1 or 0, and we only need to compare the smallest SV with the largest one to see whether the former is orders of magnitude smaller than the latter. If so, we can regard that as a zero, and proceed to retrieve the null space vector from the last column of $V^T$ (as in $A=U\Sigma V^T$) by calling Julia's {\tt svd()} function. The elements of the null space vector are the constants we look for. If not, it implies that the dimension of the null space is zero, and we conclude that the theory is invalid.

\section{Results}
\label{sec:results}

\begin{figure}[h]
\centering
\includegraphics[width=6.0truein]{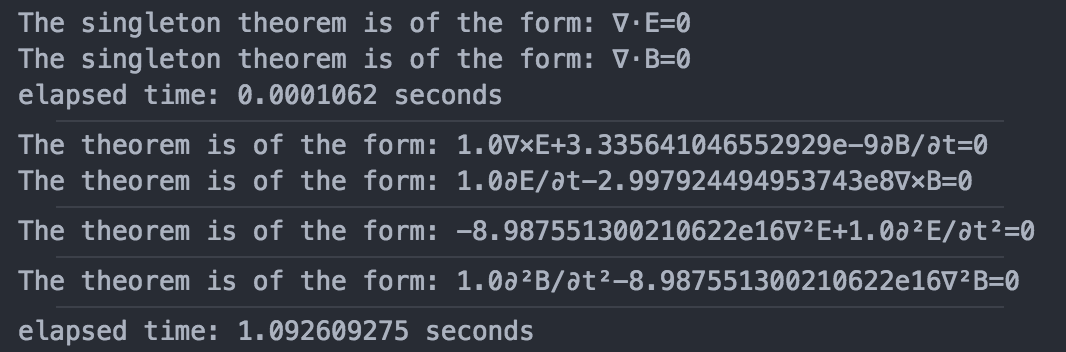}
\caption{Screenshot of \theosea's\ output showing the Maxwell Equations and the wave equations of light including the speed of light $c \approx 2.99\times 10^8.$}
\label{fig:out}
\end{figure}

\theosea\ has rediscovered the Maxwell Equations as shown in the screenshot  Fig.~\ref{fig:out}, with timings. In addition, it redisovered
\begin{eqnarray}
  {1 \over c^2} {\partial^2 \over \partial t^2} \mvec{E} - \nabb \mvec{E} & = & 0 \\
  {1 \over c^2} {\partial^2 \over \partial t^2} \mvec{B} - \nabb \mvec{B} & = & 0
\end{eqnarray}
that is a plane electromagnetic wave traveling in free space: {\em Light.}\footnote{The derivation of these wave equations from the Maxwell Equations takes humans some non-trivial vector calculus, and yet the machine did it by ``enlightened'' search.}\footnote{This machine was a MacBook Pro (Retina, 15-inch) running a 2.5 Ghz Intel Core i7 single-threaded.}

\section{Concluding Remarks and Future Work}
\label{sec:remarks}

There are many avenues for future development as briefly listed below.
\begin{itemize}
\item Generalize the language. Expand the enumeration language to allow more expressive theories. Right now \theosea\ is limited to theories of the form $c_1 A_1 + c_2 A_2 + \ldots$ where the $A$’s are operators over fields like $\mvec{E}, \mvec{B}.$ The Xu ideal gas law code works with scalar fields and exponents. Somehow these should be merged.
\item Bigger data and parallelism. The data set used was very small but semantically very rich. Other data sets, such as for macro economics, will be far larger. Here Julia's on-the-fly compilation (of candidate theories) and support for parallel processing with be very helpful, and this is one of the reasons the language was chosen\footnote{The author encourages the Julia developers to continue work on threads as that model is natural for multicore processors. For example, the main thread could enumerate candidate theories and then send them to several worker threads for validation. At any instant several theories would be under consideration.}.
\item The fully general equations can be re-discovered just by adding a current $\mvec{J}$ and source region $\rho.$ The changes to the virtual experiment and language ${\cal L}$ are straightforward.
\item Field discovery. The fields $\mvec{E, B}$ are treated as observables. It would be nice if \theosea\ could discover the fields from the forces, e.g.~$\mvec{F} = q \mvec{E}$. One step is to use a relativistic moving charge $q$ with velocity $\mvec{u},$ where the magnetic field can be written in terms of the electric field\cite{fitzpatrick08}: $\mvec{B} = (1/c^2)\,\mvec{u} {\bf \times} \mvec{E}.$ Then the field discovery problem reduces to finding the electric field and then the magnetic field will fall out from the search.
\end{itemize}
But, perhaps the most exciting extension is to apply \theosea\ to other domains such as thermodynamics, and it is beginning to look like it should work for the Schr\"{o}dinger Equation and quantum mechanics\cite{dirac67}. Work is progressing in these areas, and focusing on the applicable representation language and executable semantics are the keys for new domains.

% Back matter

\section*{Materials}

The Julia code is available with this pre-print on arXiv. The code is distributed under a Creative Commons Attribution 4.0 International Public License. If you use \theosea\ please attribute to M.A.~Stalzer and C.~Ju, \theosea: Marching Theory to Light, arXiv, August 2017.

\section*{Acknowledgements}

This research is funded by the Gordon and Betty Moore Foundation through Grant GBMF4915 to the Caltech Center for Data-Driven Discovery. Discussions with Mr.~William Xu of Caltech Math/Computer Science were very helpful. The authors are grateful to Prof.~S.G.~Djorgovski of Caltech Astronomy and Prof.~V.~Chandler of KGI Natural Sciences for their support.

% References


\begin{thebibliography}{99}

% \bibitem{maxwell73} . J.C.~Mawell {\em A Treatise on Electricity and Magnetism, Vols.~ 1\&2.} Univ.~of Oxford, Macmillan and Co., 1873. Retrieved from {\tt archive.org.}

\bibitem{stalzer16} M.~Stalzer and C.~Mentzel. A preliminary review of influential works in data-driven discovery. SpringerPlus, (5)1266, doi:10.1186/s40064-016-2888-8, 2016.

\bibitem{maxwell65} J.C.~Maxwell. A dynamical theory of the electromagnetic field. Phil.~Trans.~of the Royal Society of London, 155, 459--512, 1865.

\bibitem{korf85} R.E.~Korf. Depth-First Iterative-Deepening: An optimal admissible tree search. Artificial Intelligence, 27, 97--109, 1985.

\bibitem{langley79} P.~Langley. Rediscovering physics with BACON.3. Proc.~Intl.~Joint Conf.~on Artificial Intelligence, 505--507, 1979.

\bibitem{xu17} W.~Xu and M.~Stalzer. Deriving compact laws based on algebraic formulation of data set. arXiv:1706.05123 [cs.LG], 2017.

\bibitem{schmidt09} M.~Schmidt and H.~Lipson. Distilling free-form natural laws from experimental data. Science 324(5923), 81--85, 2009.

\bibitem{julia17} J.~Bezanson et.~al. Julia: A fresh approach to numerical computing. SIAM Review, 59(1), 65--98, doi:10.1137/141000671, 2017. See {\tt julialang.org/publications.}

%\bibitem{lorrain70} P.~Lorrain and D. Corson. {\em Electromagnetic Fields and Waves, 2nd Ed.} W.H.~Freeman, Sec.~14.1, 1970.

\bibitem{griffiths15} D.J.~Griffiths. {\em Introduction to Electrodynamics, 4th Ed.} Pearson Education, 2015.

\bibitem{denning78} P.J.~Denning, J.B.~Dennis, and J.E.~Qualitz. {\em Machines, Languages, and Computation.} Prentice-Hall, Sec.~13.8, 1978.

\bibitem{borel13} E.~Borel. M\'{e}canique statistique et irr\'{e}versibiliit\'{e}. J.~Phys., 53 s\'{e}rie, 3, 189--196, 1913.

\bibitem{stalzer17} M.~Stalzer. On the enumeration of sentences by compactness. arXiv:1706.06975 [cs:AI], 2017.

\bibitem{fitzpatrick08} R.~Fitzpatrick. {\em Maxwell's Equations and the Principles of Electromagnetism.} Firewall Media, Sec.~10.18, 2008.

\bibitem{dirac67} P.A.M.~Dirac, {\em The Principles of Quantum Mechanics.} Oxford, 1967.
\end{thebibliography}
\end{document}